\documentclass[12pt]{amsart}
\usepackage{amscd}
\usepackage{amsmath}
\usepackage{graphicx}
\usepackage{url}
\usepackage[utf8]{inputenc} 
\usepackage[T1]{fontenc}    
\usepackage[margin=34mm]{geometry}

\usepackage{hyperref}

\usepackage[draft]{minted}

\makeatletter
\def\PYGdefault@reset{\let\PYGdefault@it=\relax \let\PYGdefault@bf=\relax%
    \let\PYGdefault@ul=\relax \let\PYGdefault@tc=\relax%
    \let\PYGdefault@bc=\relax \let\PYGdefault@ff=\relax}
\def\PYGdefault@tok#1{\csname PYGdefault@tok@#1\endcsname}
\def\PYGdefault@toks#1+{\ifx\relax#1\empty\else%
    \PYGdefault@tok{#1}\expandafter\PYGdefault@toks\fi}
\def\PYGdefault@do#1{\PYGdefault@bc{\PYGdefault@tc{\PYGdefault@ul{%
    \PYGdefault@it{\PYGdefault@bf{\PYGdefault@ff{#1}}}}}}}
\def\PYGdefault#1#2{\PYGdefault@reset\PYGdefault@toks#1+\relax+\PYGdefault@do{#2}}

\expandafter\def\csname PYGdefault@tok@gd\endcsname{\def\PYGdefault@tc##1{\textcolor[rgb]{0.63,0.00,0.00}{##1}}}
\expandafter\def\csname PYGdefault@tok@gu\endcsname{\let\PYGdefault@bf=\textbf\def\PYGdefault@tc##1{\textcolor[rgb]{0.50,0.00,0.50}{##1}}}
\expandafter\def\csname PYGdefault@tok@gt\endcsname{\def\PYGdefault@tc##1{\textcolor[rgb]{0.00,0.27,0.87}{##1}}}
\expandafter\def\csname PYGdefault@tok@gs\endcsname{\let\PYGdefault@bf=\textbf}
\expandafter\def\csname PYGdefault@tok@gr\endcsname{\def\PYGdefault@tc##1{\textcolor[rgb]{1.00,0.00,0.00}{##1}}}
\expandafter\def\csname PYGdefault@tok@cm\endcsname{\let\PYGdefault@it=\textit\def\PYGdefault@tc##1{\textcolor[rgb]{0.25,0.50,0.50}{##1}}}
\expandafter\def\csname PYGdefault@tok@vg\endcsname{\def\PYGdefault@tc##1{\textcolor[rgb]{0.10,0.09,0.49}{##1}}}
\expandafter\def\csname PYGdefault@tok@vi\endcsname{\def\PYGdefault@tc##1{\textcolor[rgb]{0.10,0.09,0.49}{##1}}}
\expandafter\def\csname PYGdefault@tok@vm\endcsname{\def\PYGdefault@tc##1{\textcolor[rgb]{0.10,0.09,0.49}{##1}}}
\expandafter\def\csname PYGdefault@tok@mh\endcsname{\def\PYGdefault@tc##1{\textcolor[rgb]{0.40,0.40,0.40}{##1}}}
\expandafter\def\csname PYGdefault@tok@cs\endcsname{\let\PYGdefault@it=\textit\def\PYGdefault@tc##1{\textcolor[rgb]{0.25,0.50,0.50}{##1}}}
\expandafter\def\csname PYGdefault@tok@ge\endcsname{\let\PYGdefault@it=\textit}
\expandafter\def\csname PYGdefault@tok@vc\endcsname{\def\PYGdefault@tc##1{\textcolor[rgb]{0.10,0.09,0.49}{##1}}}
\expandafter\def\csname PYGdefault@tok@il\endcsname{\def\PYGdefault@tc##1{\textcolor[rgb]{0.40,0.40,0.40}{##1}}}
\expandafter\def\csname PYGdefault@tok@go\endcsname{\def\PYGdefault@tc##1{\textcolor[rgb]{0.53,0.53,0.53}{##1}}}
\expandafter\def\csname PYGdefault@tok@cp\endcsname{\def\PYGdefault@tc##1{\textcolor[rgb]{0.74,0.48,0.00}{##1}}}
\expandafter\def\csname PYGdefault@tok@gi\endcsname{\def\PYGdefault@tc##1{\textcolor[rgb]{0.00,0.63,0.00}{##1}}}
\expandafter\def\csname PYGdefault@tok@gh\endcsname{\let\PYGdefault@bf=\textbf\def\PYGdefault@tc##1{\textcolor[rgb]{0.00,0.00,0.50}{##1}}}
\expandafter\def\csname PYGdefault@tok@ni\endcsname{\let\PYGdefault@bf=\textbf\def\PYGdefault@tc##1{\textcolor[rgb]{0.60,0.60,0.60}{##1}}}
\expandafter\def\csname PYGdefault@tok@nl\endcsname{\def\PYGdefault@tc##1{\textcolor[rgb]{0.63,0.63,0.00}{##1}}}
\expandafter\def\csname PYGdefault@tok@nn\endcsname{\let\PYGdefault@bf=\textbf\def\PYGdefault@tc##1{\textcolor[rgb]{0.00,0.00,1.00}{##1}}}
\expandafter\def\csname PYGdefault@tok@no\endcsname{\def\PYGdefault@tc##1{\textcolor[rgb]{0.53,0.00,0.00}{##1}}}
\expandafter\def\csname PYGdefault@tok@na\endcsname{\def\PYGdefault@tc##1{\textcolor[rgb]{0.49,0.56,0.16}{##1}}}
\expandafter\def\csname PYGdefault@tok@nb\endcsname{\def\PYGdefault@tc##1{\textcolor[rgb]{0.00,0.50,0.00}{##1}}}
\expandafter\def\csname PYGdefault@tok@nc\endcsname{\let\PYGdefault@bf=\textbf\def\PYGdefault@tc##1{\textcolor[rgb]{0.00,0.00,1.00}{##1}}}
\expandafter\def\csname PYGdefault@tok@nd\endcsname{\def\PYGdefault@tc##1{\textcolor[rgb]{0.67,0.13,1.00}{##1}}}
\expandafter\def\csname PYGdefault@tok@ne\endcsname{\let\PYGdefault@bf=\textbf\def\PYGdefault@tc##1{\textcolor[rgb]{0.82,0.25,0.23}{##1}}}
\expandafter\def\csname PYGdefault@tok@nf\endcsname{\def\PYGdefault@tc##1{\textcolor[rgb]{0.00,0.00,1.00}{##1}}}
\expandafter\def\csname PYGdefault@tok@si\endcsname{\let\PYGdefault@bf=\textbf\def\PYGdefault@tc##1{\textcolor[rgb]{0.73,0.40,0.53}{##1}}}
\expandafter\def\csname PYGdefault@tok@s2\endcsname{\def\PYGdefault@tc##1{\textcolor[rgb]{0.73,0.13,0.13}{##1}}}
\expandafter\def\csname PYGdefault@tok@nt\endcsname{\let\PYGdefault@bf=\textbf\def\PYGdefault@tc##1{\textcolor[rgb]{0.00,0.50,0.00}{##1}}}
\expandafter\def\csname PYGdefault@tok@nv\endcsname{\def\PYGdefault@tc##1{\textcolor[rgb]{0.10,0.09,0.49}{##1}}}
\expandafter\def\csname PYGdefault@tok@s1\endcsname{\def\PYGdefault@tc##1{\textcolor[rgb]{0.73,0.13,0.13}{##1}}}
\expandafter\def\csname PYGdefault@tok@dl\endcsname{\def\PYGdefault@tc##1{\textcolor[rgb]{0.73,0.13,0.13}{##1}}}
\expandafter\def\csname PYGdefault@tok@ch\endcsname{\let\PYGdefault@it=\textit\def\PYGdefault@tc##1{\textcolor[rgb]{0.25,0.50,0.50}{##1}}}
\expandafter\def\csname PYGdefault@tok@m\endcsname{\def\PYGdefault@tc##1{\textcolor[rgb]{0.40,0.40,0.40}{##1}}}
\expandafter\def\csname PYGdefault@tok@gp\endcsname{\let\PYGdefault@bf=\textbf\def\PYGdefault@tc##1{\textcolor[rgb]{0.00,0.00,0.50}{##1}}}
\expandafter\def\csname PYGdefault@tok@sh\endcsname{\def\PYGdefault@tc##1{\textcolor[rgb]{0.73,0.13,0.13}{##1}}}
\expandafter\def\csname PYGdefault@tok@ow\endcsname{\let\PYGdefault@bf=\textbf\def\PYGdefault@tc##1{\textcolor[rgb]{0.67,0.13,1.00}{##1}}}
\expandafter\def\csname PYGdefault@tok@sx\endcsname{\def\PYGdefault@tc##1{\textcolor[rgb]{0.00,0.50,0.00}{##1}}}
\expandafter\def\csname PYGdefault@tok@bp\endcsname{\def\PYGdefault@tc##1{\textcolor[rgb]{0.00,0.50,0.00}{##1}}}
\expandafter\def\csname PYGdefault@tok@c1\endcsname{\let\PYGdefault@it=\textit\def\PYGdefault@tc##1{\textcolor[rgb]{0.25,0.50,0.50}{##1}}}
\expandafter\def\csname PYGdefault@tok@fm\endcsname{\def\PYGdefault@tc##1{\textcolor[rgb]{0.00,0.00,1.00}{##1}}}
\expandafter\def\csname PYGdefault@tok@o\endcsname{\def\PYGdefault@tc##1{\textcolor[rgb]{0.40,0.40,0.40}{##1}}}
\expandafter\def\csname PYGdefault@tok@kc\endcsname{\let\PYGdefault@bf=\textbf\def\PYGdefault@tc##1{\textcolor[rgb]{0.00,0.50,0.00}{##1}}}
\expandafter\def\csname PYGdefault@tok@c\endcsname{\let\PYGdefault@it=\textit\def\PYGdefault@tc##1{\textcolor[rgb]{0.25,0.50,0.50}{##1}}}
\expandafter\def\csname PYGdefault@tok@mf\endcsname{\def\PYGdefault@tc##1{\textcolor[rgb]{0.40,0.40,0.40}{##1}}}
\expandafter\def\csname PYGdefault@tok@err\endcsname{\def\PYGdefault@bc##1{\setlength{\fboxsep}{0pt}\fcolorbox[rgb]{1.00,0.00,0.00}{1,1,1}{\strut ##1}}}
\expandafter\def\csname PYGdefault@tok@mb\endcsname{\def\PYGdefault@tc##1{\textcolor[rgb]{0.40,0.40,0.40}{##1}}}
\expandafter\def\csname PYGdefault@tok@ss\endcsname{\def\PYGdefault@tc##1{\textcolor[rgb]{0.10,0.09,0.49}{##1}}}
\expandafter\def\csname PYGdefault@tok@sr\endcsname{\def\PYGdefault@tc##1{\textcolor[rgb]{0.73,0.40,0.53}{##1}}}
\expandafter\def\csname PYGdefault@tok@mo\endcsname{\def\PYGdefault@tc##1{\textcolor[rgb]{0.40,0.40,0.40}{##1}}}
\expandafter\def\csname PYGdefault@tok@kd\endcsname{\let\PYGdefault@bf=\textbf\def\PYGdefault@tc##1{\textcolor[rgb]{0.00,0.50,0.00}{##1}}}
\expandafter\def\csname PYGdefault@tok@mi\endcsname{\def\PYGdefault@tc##1{\textcolor[rgb]{0.40,0.40,0.40}{##1}}}
\expandafter\def\csname PYGdefault@tok@kn\endcsname{\let\PYGdefault@bf=\textbf\def\PYGdefault@tc##1{\textcolor[rgb]{0.00,0.50,0.00}{##1}}}
\expandafter\def\csname PYGdefault@tok@cpf\endcsname{\let\PYGdefault@it=\textit\def\PYGdefault@tc##1{\textcolor[rgb]{0.25,0.50,0.50}{##1}}}
\expandafter\def\csname PYGdefault@tok@kr\endcsname{\let\PYGdefault@bf=\textbf\def\PYGdefault@tc##1{\textcolor[rgb]{0.00,0.50,0.00}{##1}}}
\expandafter\def\csname PYGdefault@tok@s\endcsname{\def\PYGdefault@tc##1{\textcolor[rgb]{0.73,0.13,0.13}{##1}}}
\expandafter\def\csname PYGdefault@tok@kp\endcsname{\def\PYGdefault@tc##1{\textcolor[rgb]{0.00,0.50,0.00}{##1}}}
\expandafter\def\csname PYGdefault@tok@w\endcsname{\def\PYGdefault@tc##1{\textcolor[rgb]{0.73,0.73,0.73}{##1}}}
\expandafter\def\csname PYGdefault@tok@kt\endcsname{\def\PYGdefault@tc##1{\textcolor[rgb]{0.69,0.00,0.25}{##1}}}
\expandafter\def\csname PYGdefault@tok@sc\endcsname{\def\PYGdefault@tc##1{\textcolor[rgb]{0.73,0.13,0.13}{##1}}}
\expandafter\def\csname PYGdefault@tok@sb\endcsname{\def\PYGdefault@tc##1{\textcolor[rgb]{0.73,0.13,0.13}{##1}}}
\expandafter\def\csname PYGdefault@tok@sa\endcsname{\def\PYGdefault@tc##1{\textcolor[rgb]{0.73,0.13,0.13}{##1}}}
\expandafter\def\csname PYGdefault@tok@k\endcsname{\let\PYGdefault@bf=\textbf\def\PYGdefault@tc##1{\textcolor[rgb]{0.00,0.50,0.00}{##1}}}
\expandafter\def\csname PYGdefault@tok@se\endcsname{\let\PYGdefault@bf=\textbf\def\PYGdefault@tc##1{\textcolor[rgb]{0.73,0.40,0.13}{##1}}}
\expandafter\def\csname PYGdefault@tok@sd\endcsname{\let\PYGdefault@it=\textit\def\PYGdefault@tc##1{\textcolor[rgb]{0.73,0.13,0.13}{##1}}}

\makeatother

\makeatletter
\def\PYG@reset{\let\PYG@it=\relax \let\PYG@bf=\relax%
    \let\PYG@ul=\relax \let\PYG@tc=\relax%
    \let\PYG@bc=\relax \let\PYG@ff=\relax}
\def\PYG@tok#1{\csname PYG@tok@#1\endcsname}
\def\PYG@toks#1+{\ifx\relax#1\empty\else%
    \PYG@tok{#1}\expandafter\PYG@toks\fi}
\def\PYG@do#1{\PYG@bc{\PYG@tc{\PYG@ul{%
    \PYG@it{\PYG@bf{\PYG@ff{#1}}}}}}}
\def\PYG#1#2{\PYG@reset\PYG@toks#1+\relax+\PYG@do{#2}}

\expandafter\def\csname PYG@tok@gd\endcsname{\def\PYG@tc##1{\textcolor[rgb]{0.63,0.00,0.00}{##1}}}
\expandafter\def\csname PYG@tok@gu\endcsname{\let\PYG@bf=\textbf\def\PYG@tc##1{\textcolor[rgb]{0.50,0.00,0.50}{##1}}}
\expandafter\def\csname PYG@tok@gt\endcsname{\def\PYG@tc##1{\textcolor[rgb]{0.00,0.27,0.87}{##1}}}
\expandafter\def\csname PYG@tok@gs\endcsname{\let\PYG@bf=\textbf}
\expandafter\def\csname PYG@tok@gr\endcsname{\def\PYG@tc##1{\textcolor[rgb]{1.00,0.00,0.00}{##1}}}
\expandafter\def\csname PYG@tok@cm\endcsname{\let\PYG@it=\textit\def\PYG@tc##1{\textcolor[rgb]{0.25,0.50,0.50}{##1}}}
\expandafter\def\csname PYG@tok@vg\endcsname{\def\PYG@tc##1{\textcolor[rgb]{0.10,0.09,0.49}{##1}}}
\expandafter\def\csname PYG@tok@vi\endcsname{\def\PYG@tc##1{\textcolor[rgb]{0.10,0.09,0.49}{##1}}}
\expandafter\def\csname PYG@tok@vm\endcsname{\def\PYG@tc##1{\textcolor[rgb]{0.10,0.09,0.49}{##1}}}
\expandafter\def\csname PYG@tok@mh\endcsname{\def\PYG@tc##1{\textcolor[rgb]{0.40,0.40,0.40}{##1}}}
\expandafter\def\csname PYG@tok@cs\endcsname{\let\PYG@it=\textit\def\PYG@tc##1{\textcolor[rgb]{0.25,0.50,0.50}{##1}}}
\expandafter\def\csname PYG@tok@ge\endcsname{\let\PYG@it=\textit}
\expandafter\def\csname PYG@tok@vc\endcsname{\def\PYG@tc##1{\textcolor[rgb]{0.10,0.09,0.49}{##1}}}
\expandafter\def\csname PYG@tok@il\endcsname{\def\PYG@tc##1{\textcolor[rgb]{0.40,0.40,0.40}{##1}}}
\expandafter\def\csname PYG@tok@go\endcsname{\def\PYG@tc##1{\textcolor[rgb]{0.53,0.53,0.53}{##1}}}
\expandafter\def\csname PYG@tok@cp\endcsname{\def\PYG@tc##1{\textcolor[rgb]{0.74,0.48,0.00}{##1}}}
\expandafter\def\csname PYG@tok@gi\endcsname{\def\PYG@tc##1{\textcolor[rgb]{0.00,0.63,0.00}{##1}}}
\expandafter\def\csname PYG@tok@gh\endcsname{\let\PYG@bf=\textbf\def\PYG@tc##1{\textcolor[rgb]{0.00,0.00,0.50}{##1}}}
\expandafter\def\csname PYG@tok@ni\endcsname{\let\PYG@bf=\textbf\def\PYG@tc##1{\textcolor[rgb]{0.60,0.60,0.60}{##1}}}
\expandafter\def\csname PYG@tok@nl\endcsname{\def\PYG@tc##1{\textcolor[rgb]{0.63,0.63,0.00}{##1}}}
\expandafter\def\csname PYG@tok@nn\endcsname{\let\PYG@bf=\textbf\def\PYG@tc##1{\textcolor[rgb]{0.00,0.00,1.00}{##1}}}
\expandafter\def\csname PYG@tok@no\endcsname{\def\PYG@tc##1{\textcolor[rgb]{0.53,0.00,0.00}{##1}}}
\expandafter\def\csname PYG@tok@na\endcsname{\def\PYG@tc##1{\textcolor[rgb]{0.49,0.56,0.16}{##1}}}
\expandafter\def\csname PYG@tok@nb\endcsname{\def\PYG@tc##1{\textcolor[rgb]{0.00,0.50,0.00}{##1}}}
\expandafter\def\csname PYG@tok@nc\endcsname{\let\PYG@bf=\textbf\def\PYG@tc##1{\textcolor[rgb]{0.00,0.00,1.00}{##1}}}
\expandafter\def\csname PYG@tok@nd\endcsname{\def\PYG@tc##1{\textcolor[rgb]{0.67,0.13,1.00}{##1}}}
\expandafter\def\csname PYG@tok@ne\endcsname{\let\PYG@bf=\textbf\def\PYG@tc##1{\textcolor[rgb]{0.82,0.25,0.23}{##1}}}
\expandafter\def\csname PYG@tok@nf\endcsname{\def\PYG@tc##1{\textcolor[rgb]{0.00,0.00,1.00}{##1}}}
\expandafter\def\csname PYG@tok@si\endcsname{\let\PYG@bf=\textbf\def\PYG@tc##1{\textcolor[rgb]{0.73,0.40,0.53}{##1}}}
\expandafter\def\csname PYG@tok@s2\endcsname{\def\PYG@tc##1{\textcolor[rgb]{0.73,0.13,0.13}{##1}}}
\expandafter\def\csname PYG@tok@nt\endcsname{\let\PYG@bf=\textbf\def\PYG@tc##1{\textcolor[rgb]{0.00,0.50,0.00}{##1}}}
\expandafter\def\csname PYG@tok@nv\endcsname{\def\PYG@tc##1{\textcolor[rgb]{0.10,0.09,0.49}{##1}}}
\expandafter\def\csname PYG@tok@s1\endcsname{\def\PYG@tc##1{\textcolor[rgb]{0.73,0.13,0.13}{##1}}}
\expandafter\def\csname PYG@tok@dl\endcsname{\def\PYG@tc##1{\textcolor[rgb]{0.73,0.13,0.13}{##1}}}
\expandafter\def\csname PYG@tok@ch\endcsname{\let\PYG@it=\textit\def\PYG@tc##1{\textcolor[rgb]{0.25,0.50,0.50}{##1}}}
\expandafter\def\csname PYG@tok@m\endcsname{\def\PYG@tc##1{\textcolor[rgb]{0.40,0.40,0.40}{##1}}}
\expandafter\def\csname PYG@tok@gp\endcsname{\let\PYG@bf=\textbf\def\PYG@tc##1{\textcolor[rgb]{0.00,0.00,0.50}{##1}}}
\expandafter\def\csname PYG@tok@sh\endcsname{\def\PYG@tc##1{\textcolor[rgb]{0.73,0.13,0.13}{##1}}}
\expandafter\def\csname PYG@tok@ow\endcsname{\let\PYG@bf=\textbf\def\PYG@tc##1{\textcolor[rgb]{0.67,0.13,1.00}{##1}}}
\expandafter\def\csname PYG@tok@sx\endcsname{\def\PYG@tc##1{\textcolor[rgb]{0.00,0.50,0.00}{##1}}}
\expandafter\def\csname PYG@tok@bp\endcsname{\def\PYG@tc##1{\textcolor[rgb]{0.00,0.50,0.00}{##1}}}
\expandafter\def\csname PYG@tok@c1\endcsname{\let\PYG@it=\textit\def\PYG@tc##1{\textcolor[rgb]{0.25,0.50,0.50}{##1}}}
\expandafter\def\csname PYG@tok@fm\endcsname{\def\PYG@tc##1{\textcolor[rgb]{0.00,0.00,1.00}{##1}}}
\expandafter\def\csname PYG@tok@o\endcsname{\def\PYG@tc##1{\textcolor[rgb]{0.40,0.40,0.40}{##1}}}
\expandafter\def\csname PYG@tok@kc\endcsname{\let\PYG@bf=\textbf\def\PYG@tc##1{\textcolor[rgb]{0.00,0.50,0.00}{##1}}}
\expandafter\def\csname PYG@tok@c\endcsname{\let\PYG@it=\textit\def\PYG@tc##1{\textcolor[rgb]{0.25,0.50,0.50}{##1}}}
\expandafter\def\csname PYG@tok@mf\endcsname{\def\PYG@tc##1{\textcolor[rgb]{0.40,0.40,0.40}{##1}}}
\expandafter\def\csname PYG@tok@err\endcsname{\def\PYG@bc##1{\setlength{\fboxsep}{0pt}\fcolorbox[rgb]{1.00,0.00,0.00}{1,1,1}{\strut ##1}}}
\expandafter\def\csname PYG@tok@mb\endcsname{\def\PYG@tc##1{\textcolor[rgb]{0.40,0.40,0.40}{##1}}}
\expandafter\def\csname PYG@tok@ss\endcsname{\def\PYG@tc##1{\textcolor[rgb]{0.10,0.09,0.49}{##1}}}
\expandafter\def\csname PYG@tok@sr\endcsname{\def\PYG@tc##1{\textcolor[rgb]{0.73,0.40,0.53}{##1}}}
\expandafter\def\csname PYG@tok@mo\endcsname{\def\PYG@tc##1{\textcolor[rgb]{0.40,0.40,0.40}{##1}}}
\expandafter\def\csname PYG@tok@kd\endcsname{\let\PYG@bf=\textbf\def\PYG@tc##1{\textcolor[rgb]{0.00,0.50,0.00}{##1}}}
\expandafter\def\csname PYG@tok@mi\endcsname{\def\PYG@tc##1{\textcolor[rgb]{0.40,0.40,0.40}{##1}}}
\expandafter\def\csname PYG@tok@kn\endcsname{\let\PYG@bf=\textbf\def\PYG@tc##1{\textcolor[rgb]{0.00,0.50,0.00}{##1}}}
\expandafter\def\csname PYG@tok@cpf\endcsname{\let\PYG@it=\textit\def\PYG@tc##1{\textcolor[rgb]{0.25,0.50,0.50}{##1}}}
\expandafter\def\csname PYG@tok@kr\endcsname{\let\PYG@bf=\textbf\def\PYG@tc##1{\textcolor[rgb]{0.00,0.50,0.00}{##1}}}
\expandafter\def\csname PYG@tok@s\endcsname{\def\PYG@tc##1{\textcolor[rgb]{0.73,0.13,0.13}{##1}}}
\expandafter\def\csname PYG@tok@kp\endcsname{\def\PYG@tc##1{\textcolor[rgb]{0.00,0.50,0.00}{##1}}}
\expandafter\def\csname PYG@tok@w\endcsname{\def\PYG@tc##1{\textcolor[rgb]{0.73,0.73,0.73}{##1}}}
\expandafter\def\csname PYG@tok@kt\endcsname{\def\PYG@tc##1{\textcolor[rgb]{0.69,0.00,0.25}{##1}}}
\expandafter\def\csname PYG@tok@sc\endcsname{\def\PYG@tc##1{\textcolor[rgb]{0.73,0.13,0.13}{##1}}}
\expandafter\def\csname PYG@tok@sb\endcsname{\def\PYG@tc##1{\textcolor[rgb]{0.73,0.13,0.13}{##1}}}
\expandafter\def\csname PYG@tok@sa\endcsname{\def\PYG@tc##1{\textcolor[rgb]{0.73,0.13,0.13}{##1}}}
\expandafter\def\csname PYG@tok@k\endcsname{\let\PYG@bf=\textbf\def\PYG@tc##1{\textcolor[rgb]{0.00,0.50,0.00}{##1}}}
\expandafter\def\csname PYG@tok@se\endcsname{\let\PYG@bf=\textbf\def\PYG@tc##1{\textcolor[rgb]{0.73,0.40,0.13}{##1}}}
\expandafter\def\csname PYG@tok@sd\endcsname{\let\PYG@it=\textit\def\PYG@tc##1{\textcolor[rgb]{0.73,0.13,0.13}{##1}}}

\makeatother

\begin{document}
\bibliographystyle{plainurl}
\title[Flexible model composition]%
{Flexible model composition in machine learning\\
  and its implementation in MLJ}

\author{Anthony D.~Blaom${}^\dagger$ and Sebastian J.~Vollmer${}^\star$}
\address{Alan Turing Institute, ${}^\dagger$University of Auckland,
  ${}^\star$University of Warwick}
\thispagestyle{empty}
\begin{abstract}
  A graph-based protocol called `learning networks' which combine
  assorted machine learning models into meta-models is
  described. Learning networks are shown to overcome several
  limitations of model composition as implemented in the dominant
  machine learning platforms. After illustrating the protocol in
  simple examples, a concise syntax for specifying a learning network,
  implemented in the \texttt{MLJ} framework, is presented. Using the
  syntax, it is shown that learning networks are are sufficiently
  flexible to include Wolpert's model stacking, with out-of-sample
  predictions for the base learners.
\end{abstract}
\maketitle

\section{Introduction}

This paper details a general scheme for composing machine learning
models, implemented in the open-source machine learning toolbox
\texttt{MLJ} (Machine Learning in Julia) but likely to be of interest
more generally \cite{MLJDocs}. The paper \cite{BlaomEtal2020} gives an
overview of \texttt{MLJ}'s design without providing the detail on
model composition, which is this paper's exclusive focus.

An increasingly essential feature of a machine learning toolbox is a
facility for combining basic machine learning elements into more
sophisticated meta-models. The earliest example of such model
composition is the simple non-branching pipeline. A pipeline model
typically combines, in sequence, several pre-processing operations ---
such as type coercion and missing-value imputation --- with a final
supervised learning model.  It seems pipelines were first popularized
by the \texttt{scikit-learn} toolbox
\cite{Pedregosa2001,Buitinck2013}.

Another early example of a composite model is the homogeneous
ensemble. Here the predictions of a large number of simple atomic
learners are aggregated; each is trained with the same algorithm and
each learner shares the same hyperparameters, but incorporates some
random element to increase variance. The most well-known example of
this is the random forest, whose atomic elements are decision trees
\cite{Breiman2001}.

Inhomogeneous ensembles, blending the predictions of a relatively
small number of different but sophisticated models has also been shown
to improve performance. The most advanced model composition of this
kind is known as \textsl{model stacking} \cite{Wolpert1992} and is
used routinely by winning teams of data science competitions, such as
kaggle.

However, several limitations surrounding model composition are
increasingly evident to users of the dominant machine learning
software platforms, which were not developed with flexible model
composition in mind. For instance, the basic model composition
interfaces provided by \texttt{mlr}
\cite{BischlEtal2016}, \texttt{caret} \cite{Kuhn2008},
\texttt{sckit-learn} \cite{Pedregosa2001,Buitinck2013}, and
\texttt{Weka} \cite{Holmes1994} all share one or more of the following
shortcomings:
\begin{enumerate}
\item Composite models do not inherit all the behavior of ordinary
  models.
\item Composition is limited to linear (non-branching) pipelines.
\item Supervised components in a linear pipeline can only occur at the
  end of the pipeline.
\item Only static (unlearned) target transformations / inverse
  transformations are supported.\label{four}
\item Hyper-parameters in homogeneous model ensembles cannot be coupled.
\item Some sophisticated inhomogeneous ensembling, such as stacking (with
  out-of-sample predictions for base learners) cannot be implemented.
\item Composite models cannot implement multiple operations, for example,
  both a `predict' and `transform' method (as in clustering models) or
  both a `transform' and `inverse transform' method.
\end{enumerate}

The purpose of this article is to: (i) outline a model composition
scheme flexible enough to mitigate the above shortcomings; and (ii)
describe a model composition syntax adopted in \texttt{MLJ} which
implements the proposed scheme.

The design of \texttt{MLJ} is partly inspired by that of \texttt{mlr}.
The latter package's re-incarnation \texttt{mlr3}, developed
concurrently, also overcomes the obstacles mentioned above, with
exception of (\ref{four}) \cite{Lang2019}. The approach taken there is
not the same, however.

\subsection{Sources of current design limitations}

Composite machine learning models are generally conceived as some kind
of directed acyclic graph structure $G$, whose nodes are the component
models. In our assessment, existing design limitations arise from two
common design decisions:

Firstly, a given node is associated simultaneously with a unique set of
hyper-parameters, a corresponding set of learned parameters, and a single
operation, such as `predict' or `transform'.  However, in the case
of pipeline target transformations, for example, you want two nodes
with different operations (transform and inverse transform) but which
point to the same learned parameters. In a homogeneous ensemble, you
want multiple nodes pointing to the same model hyper-parameters, but
enjoying distinct learned parameters.

Secondly, there may be an implicit requirement that the same graph
structure essentially reflect both the flow of information during
training as in prediction.  An example where this is too restrictive
is model stacking \cite{Wolpert1992}, where: (i) each base learner
computes multiple sets of learned parameters, one set for each fold of
the provided data, to obtain an out-of-sample base-learner prediction,
used to train the adjudicator; but (ii) the adjudicating model looks
to base model nodes trained on {\textit all} the training data when
predicting on new data.

\subsection{The main idea}
In our conception a {\slshape node} is just some object that can be
called upon to deliver data, lazily computed in some way. To separate
the various objects conflated in existing designs, it is convenient to
introduce one mild abstraction, which we call a {\slshape machine}. A
machine is an object \textit{pointing to} a set of model hyperparameters $h$
(the {\slshape model}) and a sequence of nodes $N_1, N_2, ..., N_k$
(the {\slshape arguments}) from a base graph $G$. These nodes indicate
where the model should look for its training data. In training, a
single set of learned parameters is associated with each machine.

A composite model is then specified by specifying the underlying graph
$G$ and by {\itshape labeling} certain nodes --- called {\slshape
  dynamic} --- with machines and corresponding operations (such as
predict or inverse transform), and certain other nodes --- called
{\slshape static} --- with ordinary functions.  Training a dynamic
node means training the machine that labels it (i.e., training the
model specified by the machine). To train the composite model as a
whole, machines must be individually trained in an appropriate
order. The labeled graph is called a \textsl{learning network}.

We re-iterate that the learned parameters of training a machine {\itshape
  are always associated with the machine} and not with any particular
node or model. Two nodes can be labeled with same machine (but
different operations) and we allow distinction between two machines
with identical model and node specification.

Our idea is clarified in a simple example presented in \ref{simple}.

\vspace{\baselineskip} Learning networks are described in Section
\ref{formal}, which includes some simple examples. Section
\ref{section3} introduces a concise syntax for specifying a learning
network, implemented in \texttt{MLJ}, and goes on to show how
Wolpert's model stacking can be implemented.

\subsection*{Acknowledgements}
Seed funding for the \texttt{MLJ} project has been provided by the
Alan Turing Institute's Tools, Practices and Systems programme. The
authors are indebted to \texttt{MLJ} collaborators Franz
Kir\'aly, Thibaut Lienart and Diego Arenas for helpful feedback on
model composition design.

\section{A formal specification for model composition}\label{formal}

\subsection{Learners}

To train a decision tree requires the specification of
hyper-parameters, such as the maximum tree depth. Such a specification
(algorithm + hyper-parameters) is here called a {\slshape model} (also
known as a {\slshape hypothesis} or {\slshape learning strategy}). The
learning algorithm itself will be understood as a family of {\slshape
  fitting functions}\footnote{Even `non-deterministic' algorithms,
  such as random forests, can be viewed as functions if we regard the
  random number generator seed as a hyper-parameter.}
$(X, y) \mapsto f_h(X, y) \in \Theta$, one for each model $h$. Here
$\Theta$ is the space of all possible decision trees, which we more
generally refer to as {\slshape learned parameters}. Here $X$ and $y$
are feature observations and target observations for training the
decision tree.

Once a learned parameter (tree) $\theta = f_h(X, y)$ is computed,
predictions for new feature observations $X'$ are given by
$p(\theta, X')$, for some function $p$, the prediction {\slshape
  operation}.

A decision tree is an example of a {\slshape learner}. More generally, we
are interested in forming new learners by combining (composing) a
number of existing learners, where, in general, a {\slshape learner}
consists of:

\begin{enumerate}
\item a set $H$ of {\slshape models}, each trained using the using the same algorithm
\item a set $\Theta$ of {\slshape learned parameters}
\item {\slshape fitting functions}
  $({\mathbf X}_1, {\mathbf X}_2, \ldots, {\mathbf X_{k_H}}) \mapsto
  f_h({\mathbf X}_1, {\mathbf X_2}, \ldots, {\mathbf X}_{k_H}) \in
  \Theta$, one for each $h \in H$; and
\item one or more {\slshape operations} of the form
  $p \colon \Theta \times {\mathcal X_p} \rightarrow {\mathcal Y_p} $
\end{enumerate}
The variables
${\mathbf X}_1, {\mathbf X}_2, \ldots, {\mathbf X}_{k_H}$ are called
{\slshape training arguments}.

The learner is {\slshape unsupervised} if $k_H = 1$ and there exists an
operation
$t \colon \Theta \times {\mathcal X_t} \rightarrow {\mathcal Y_t} $,
called the {\slshape transformation} operation, subject to the
understanding that $X_1$ is always drawn from ${\mathcal X}_t$. Some
unsupervised learners will have an {\slshape inversion} operation $i$, with
the property that $i(\theta, \, \cdot \,)$ and
$t(\theta, \, \cdot \,)$ are inverses (or approximately so).

The learner is {\slshape supervised} if $k_H \ge 2$ and there exists an
operation
$p\colon \Theta \times {\mathcal X_p} \rightarrow {\mathcal Y_p} $
called the {\slshape prediction} operation such that $X_1$ is drawn
from ${\mathcal X}_p$ and $X_2$ from ${\mathcal Y}_p$. (A third
argument $X_3$ of the fitting functions might represent, for example,
sample weights, for supervised learners that support them.)

\subsection{A simple example of a learning network}\label{simple}

A {\slshape composite learner} is one whose fitting functions and
operations are encoded in a certain directed graph, here called a {\slshape
  learning network}, labeled with metadata pertaining to the learners to
be composed.

Before stating the definition of learning networks, we informally
describe the example illustrated in Figure 1. 
This combines an ordinary supervised
learner $h_{\mathrm S}$, with prediction $p_{\mathrm S}$, and a
learned transformation of the target $h_{\mathrm T}$, with
transformation $t_{\mathrm T}$ and inverse $i_{\mathrm
  T}$. Specifically: (i) The network learns a target transformation
(such as normalization) using training data supplied at $y$; (ii) the
supervised model is trained using features supplied at $X$, together
with the transformed training target fetched from $z$; (iii) the
network outputs at $\hat z$ the predictions of the supervised model,
on new features to be supplied at $X$; and (iv) applies the inverse of
the target transformation learned in (i) to the predictions at
$\hat z$ to obtain the final output $\hat y$ (restoring the original
target scale, in the example of normalization).

\begin{figure}[h]\label{target_transformation}

  \includegraphics[scale=0.5]{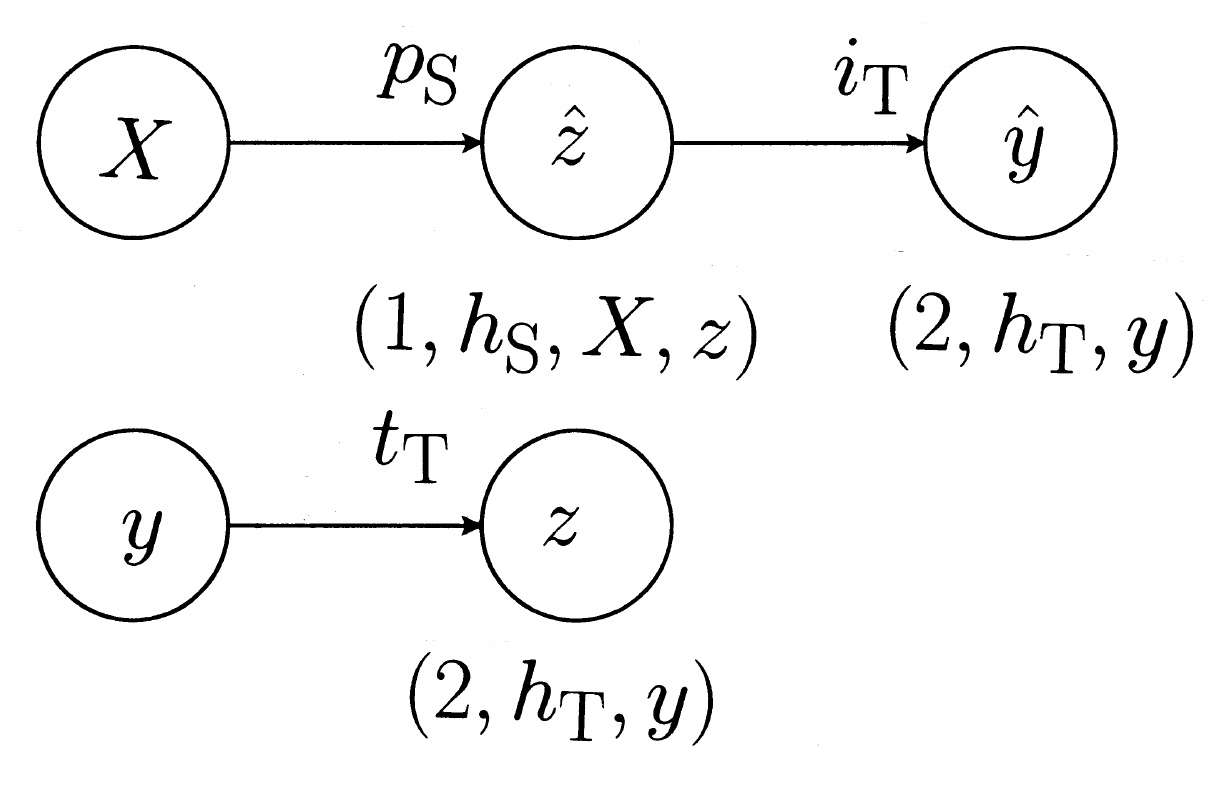}
  \caption{A learning network combining a supervised model
    $h_{\mathrm S}$ with target transformation $h_{\mathrm T}$. The
    supervised composite model receives training features and training
    target at $X$ and $y$; when new features are supplied at $X$,
    predictions are fetched from $\hat y$. Nodes $z$ and $\hat y$ are
    labeled with the same machine, corresponding to a single learned
    transformation of the target.}
\end{figure}

Observe that each non-source node is labeled with both an operation
--- $p_{\mathrm S}$, $i_{\mathrm T}$, or $t_{\mathrm T}$ --- and
tuple, such as $(2, h_{\mathrm T}, y)$, called a {\slshape
  machine}. The operation indicates how the node should process
incoming information, while the machine refers to a training event on
which the operation depends. For example, the node $\hat y$ applies
the inverse transform $i_{\mathrm T} $ to incoming data, using the
parameter $\theta $ learned by fitting the model $h_{\mathrm T}$ to
data fetched from the source node $y$. The node $z$ similarly depends
on the same training event (machine) and so receives the same machine
as label. The first number in the machine is purely an
identifier. (Two machines could specify the same model and nodes, but
would be distinct, and hence be associated with distinct learned
parameters, if their identifiers are different.)

\subsection{Machines and learning networks defined}\label{mach}

Let $G$ be a graph. Then a {\slshape machine} over $G$ is any tuple of the
  form $m = (i, h, N_1, N_2, \ldots, N_k) $, where $i$ is an integer
  (the {\slshape identifier}), $h$ is a model associated with some learner
  $L$ having $k$ training arguments, and $N_1$, $N_2$, \ldots, $N_k$
  are nodes of $G$ (the machine {\slshape training arguments}). We call $L$
  the {\slshape learner of $m$}. The learning network above has exactly two
  machines (one of which happens to label two different nodes).

Formally, a {\slshape learning network} consists of:

\begin{enumerate}
\item a finite, directed, simple, acyclic graph $G$, subject to the
  restriction that each connected component of $G$ has a unique source
  node
\item an enumeration of the set of incoming edges of any non-source
  node $N$ of $G$, or, equivalently, an enumeration
  $N^1, N^2, \ldots, N^{k_N}$ of those nodes with an outgoing edge
  ending at $N$
\item an enumeration $S^1, S^2, \ldots, S^{k_S}$ of the set of source
  nodes
\item a declaration of each non-source node of $N \in G$ as {\slshape
    static} or {\slshape dynamic} such that dynamic nodes have unique
  incoming edges
\item a labeling of each static node $N \in G$ with a function
  $(X_1, X_2,\dots, X_{k_N}) \mapsto p_N(X_1,X_2,\ldots, X_{k_N}) $,
  where $k_N$ is the number of incoming edges
\item a finite sequence of machines $m_1, m_2, \ldots, m_R$ of
  machines over $G$, such that $m_j$ has identifier $j$
\item a labeling of each dynamic node $N \in G$ with: (i) an integer
  $j$ representing the identifier of one of the machines; and (ii) and
  an operation $p_N$ for the learner of the machine labeling $N$.
\item for each operation $p$ that the corresponding composite learner
  is to support, a declaration of some node $N(p)$.\label{last}

\end{enumerate}

Elaborating on the last requirement, if, for example, the composite
model defined by the network is be considered a supervised model, with
a single predict operation, then the corresponding node indicates
where predictions are to be fetched.

Note that the same machine may label distinct dynamic nodes, and two
machines may specify the same model and/or training nodes.

All non-source nodes in the simple target transformation example of
the previous section (Figure 1) 
have a single incoming edge, but there is no static node.

The learning network shown in Figure 2 
specifies a homogeneous ensemble of a three supervised learners whose
predictions are to be aggregated.  Each dynamic node $\hat y_1$,
$\hat y_2$ and $\hat y_3$ is labeled with a separate machine (and so
will predict using a separate learned parameter). However, as each
machine specifies the same model $h$, the model hyper-parameters are
coupled (as the tree parameters in a random forest). The node $\hat y$
is static and its static operation (function) $\mu$ represents
aggregation (e.g., ``compute mean'' in the case $h$ is a regressor).

\begin{figure}[h]\label{homogeneous_ensemble}
  \centering
  \includegraphics[scale=0.5]{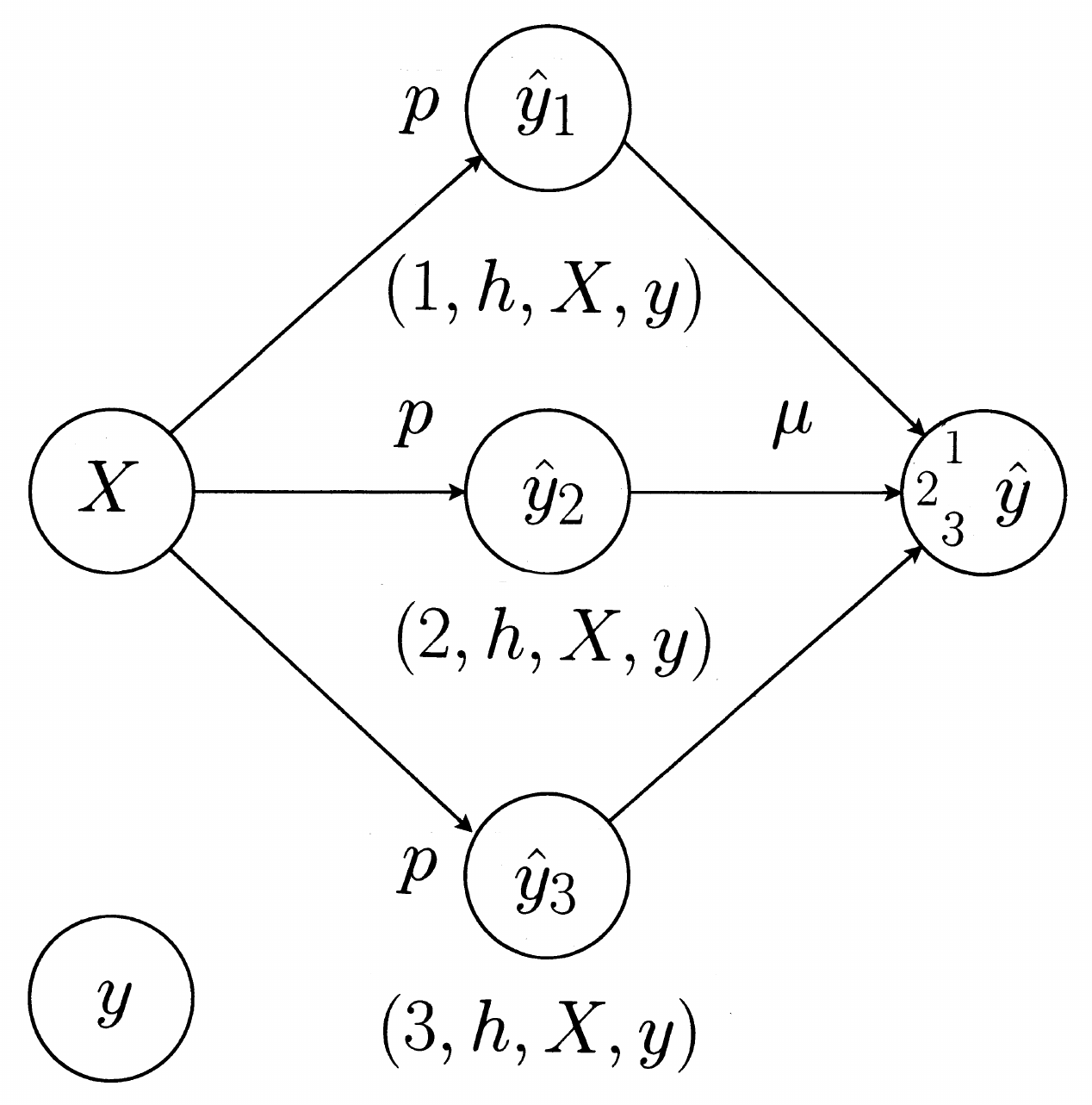}
  \caption{A three-model homogeneous ensemble with atomic model $h$,
    which is shared by the three distinct machines labeling the nodes
    $\hat y_1$, $\hat y_2$, $\hat y_3$. Predictions are aggregated
    using the function $\mu$ at node $\hat y$.}
\end{figure}

A learning network for model stacking is described in Section
\ref{section3}.

\subsection{The completion of a learning network} The underlying graph
$G$ in any learning network may be enlarged to a graph $\bar G$, here
called the {\slshape completion} of $G$, by adding a {\slshape
  training edge} between nodes $N$ and $J$ whenever $N$ appears as a
training argument of a machine labeling $J$. The training edges in
learning network of Figure 2
are shown as dashed arrows in Figure 3.
The complication of the completed network, even in this simple and
common use-case, highlights the subtlety of model composition in
machine learning.

\begin{figure}[h]\label{training_edges}
  \centering
  \includegraphics[scale=0.5]{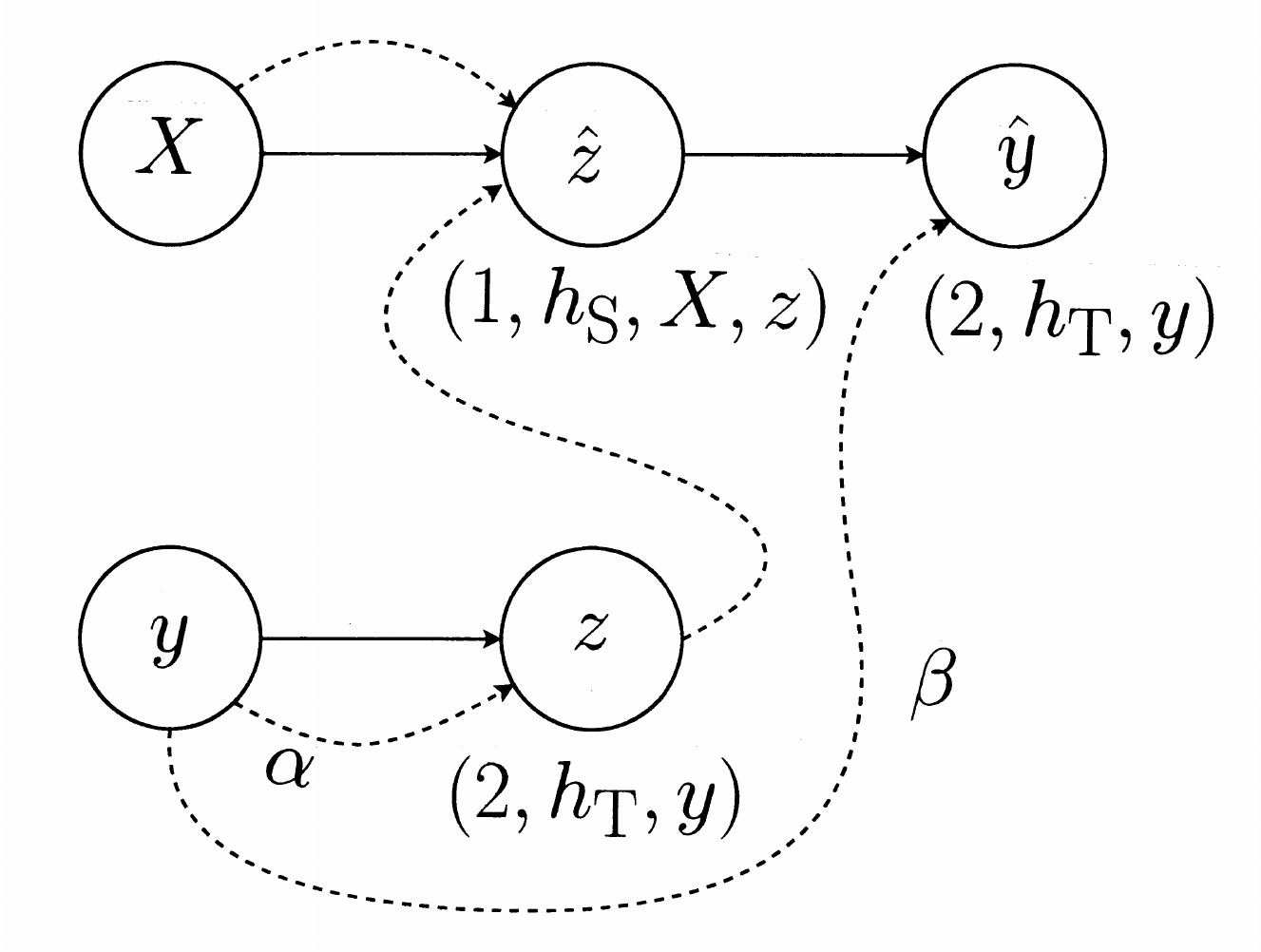}
  \caption{The learning network of Figure \ref{target_transformation}
    with training edges added (shown as dotted arrows).}
\end{figure}

When one applies an operation to a learning network --- for example,
calls on the supervised learning network in Figure 3 
to make a prediction --- then the training edges
play no role. However, as we detail in \ref{grut}, training a learning
network means training it's machines, and training a machine means
calling on its training arguments to deliver training data. In this
case one may imagine data flowing along the training edges, with the
following caveat: Since the same machine may label multiple nodes ($z$
and $\hat y$ in the figure are labeled by the same machine) distinct
training edges may actually represent the same data flow (e.g.,
$\alpha$ and $\beta$).

The main significance of the graph $\bar G$ is that encodes
dependencies. A machine $m$ labeling a node $N$ is ready to be
trained when all machines labeling nodes upstream of $N$ in $\bar G$
are already trained.

\subsection{The composite learner defined by a learning network}\label{grut}
The magnanimous reader will likely guess at the manner in which a
learning network $G$ can be interpreted as a new composite learner $L$
without detailed explanation. However, for completeness, and to
mitigate any ambiguity, we provide details below.

Our main goal is to describe the fitting function of the composite
learner. For conceptual simplicity, we describe this function in a
rather naive way; in the \texttt{MLJ} implementation component models
are trained asynchronously but this detail is not addressed here.

After removing duplicates, we can associate with the machines
$m_1, m_2, \ldots, m_R$ of the learning network $G$, a possibly
shorter sequence of unique models $h_1, h_2, \ldots h_r$, drawn from
sets of models $H_1, H_2, \ldots, H_r$.  By definition, the set of
models associated with $L$ is the Cartesian product
$H = H_1\times H_2 \times \cdots \times H_r$. The set of learned
parameters of $L$ is defined by
$\Theta = \Theta_1 \times \Theta_2 \times \cdots \Theta_R$, where
$\Theta_i$ is the set of learned parameters for the learner associated
with machine $m_i$.  So while there may be fewer factors in the set of
models than machines, there is exactly one factor in the space of
learned parameters for each machine.

We now describe the fitting function for $L$.

Suppose we are given training data $X_1, X_2, \ldots, X_k$, where $k$
is the number of source nodes. We will simultaneously define learned
the parameters $\theta_i \in \Theta_i$ to be associated with each
machine $m_i$, and data, denoted $N()$, to be associated with the
output of each node $N$ of $G$.

First, let $N_1, N_2, \ldots, N_K$ be any enumeration of the nodes of
$G$ consistent with the partial ordering of the directed acyclic graph
$\bar G$. That is, if $N_i$ is upstream of $N_j$ in $\bar G$, then
$ i < j$. From the list of corresponding machine labels, drop any
machine with an identical predecessor, obtaining a unique list of
machines. Relabeling if necessary, we may assume, without loss of
generality, that this list is precisely $m_1, m_2, \ldots , m_R$. This
ordering of machines ensures that the following definition of
$\theta_j$, inductive on index $j$, is valid:
\begin{equation*}
  \theta_j = f_{h_j}(N_1^j(), N_2^j(), \ldots, N_{k^j}^j()),
\end{equation*}
where $h_j$ is the model associated with $m_j$,
$N_1^j, N_2^j, \ldots, N_{k^j}^j$ are the training arguments of $m_j$,
and where the definition of $N()$, for each
$N \in \{N_1^j, N_2^j, \ldots, N_{k^j}^j\}$, depends on whether $N$ is
dynamic, static, or a source node, as follows.

If $N$ is dynamic, then

\begin{equation}
  N() = p_N(J(), \mu),\label{eq1}
\end{equation}
where $J$ is the (unique) input node of $N$, and $\mu$ the learned
parameter associated with the machine labeling $N$. If $N$ is static,
then instead,
\begin{equation}
  N() = p_N(J_1(), J_2(), \ldots, J_l()),\label{eq2}
\end{equation}
where $J_1, J_2, \ldots J_l$ are the input nodes of $N$.  If $J$ is a
source node, then
\begin{equation}
  N() = X_i,\label{eq3}
\end{equation}
where $i$ is the index of $N$ in the prescribed enumeration of source
nodes.

With $\theta_1, \theta_2, \ldots, \theta_R$ so defined, the fitting
function for $L$ is

$$f_{(h_1, \ldots, h_r)}(X_1, X_2, \ldots, X_k) = (\theta_1,
\theta_2, \ldots, \theta_R)$$.

Having defined training, let us finish by explaining how the network
operates in `prediction' mode.  That is, we need to define the map
$p \colon \Theta \times {\mathcal X_p} \rightarrow {\mathcal Y_p}$ for
each operation $p$ to be defined for the composite model. To this end,
we extend the recursive definitions \eqref{eq1}--\eqref{eq3} to allow
calling a node with a single argument $X'$:

\begin{align*}
  N(X') &= p_N(J(X'), \mu) \enspace &&\text{if $N$ is dynamic}\\
  N(X') &= p_N(J_1(X'), J_2(X'), \ldots, J_l(X')) \enspace &&\text{if $N$ is static}\\
  N(X') &= X'\label{eq3} \enspace &&\text{if $N$ is a source node.}
\end{align*}
Then the new operations are defined by
$p(\theta_1, \ldots, \theta_R, X') = N(p)(X')$, where $N(p)$ is the
node declared in condition \ref{mach}\eqref{last}.


\section{An advanced application: model stacking}\label{section3}

We now introduce the syntax used in \texttt{MLJ} for defining a
learning network, and demonstrate show how model stacking can be
implemented using such a syntax. A pictorial description of the
underlying graph is probably too complicated here to be immediately
useful, and we do not attempt to provide one.

Because Julia foregoes abstractions such as classes in favor of a more
functional paradigm, this syntax is close to the mathematical
description given already, and familiarity with Julia will not be
necessary to understand what follows.

\subsection{A syntax for learning networks}

Recall once again that a \textsl{model} is just a set of
hyperparameters for some machine learning algorithm (in \texttt{MLJ},
an instance of a composite type). A machine, as defined in \ref{mach},
is an object pointing to some model and some training arguments
(nodes), and is constructed in MLJ with the syntax

\begin{minted}[fontsize=\small]{julia}
    mach =  machine(model, arg1, arg2, ...)
\end{minted}

\noindent This object, which is mutable, additionally stores learned parameters
after training.\footnote{In Julia the mutability has the consequence that the
machine identifiers referred to earlier are redundant --- a second
machine constructed with the same call will be considered a distinct
object.}

A source node is constructed with the syntax \texttt{X = source()} and
a new node is generated in one of two ways:

\begin{enumerate}
\item To generate a \textit{dynamic} node labeled with machine
  \texttt{mach} and a given operation --- such as \texttt{predict} or
  \texttt{transform} --- we \textit{call} the operation on the machine
  and the node that is to be the parent of that node (always
  unique). For example, we might declare \texttt{yhat = predict(mach,
    X)}. In this case, when called on to do so, the new node
  \texttt{yhat} fetches feature data from node \texttt{X} and
  determines the prediction using learned parameters stored at
  \texttt{mach}.\label{first}
\item To generate a \textit{static} node, a function that ordinarily
  takes data as arguments (vectors, tables, etc) is simply overloaded
  to act on nodes, with a node as return value. In \texttt{MLJ} this
  can be done using a macro, but here we shall tacitly assume all
  functions have already been overloaded in this way. So, for example,
  if \texttt{y1} and \texttt{y2} are two nodes for delivering
  equi-length vector data, then \texttt{y1 + y2} is a new static node
  which, when called to do so, fetches data from nodes \texttt{y1} and
  \texttt{y2} and adds the result. Similarly, \texttt{z = mean(y1)}
  defines a new node \textit{z} that computes the mean of data fetched
  from \texttt{y1}.
\end{enumerate}

To illustrate \eqref{first}, consider again the learning network
depicted in Figure 2,
for wrapping a supervised learner in a target transformation / inverse
transformation. For concreteness, suppose the supervised learner $h_S$
is a decision tree regressor, and the target transformer $h_T$ is a
standardizer (whitener). Then the following is valid \texttt{MLJ} code
defining the composite learning network:

\begin{minted}[frame=lines,mathescape=true,escapeinside=||,fontsize=\small]{julia}
# load necessary libraries:
using MLJ
@load DecisionTreeRegressor

# instantiate model instances:
tree = DecisionTreeRegressor(min_samples_split=5)
standardizer = Standardizer()

X = source()
y = source()

# the machine (2, h_T, y):
mach2 = machine(standardizer, y)

z = transform(mach2, y)

# the machine (1, h_S, X, z):
mach1 = machine(tree, X, z)

zhat = predict(mach1, X)
yhat = inverse_transform(mach2, zhat)
\end{minted}

In \texttt{MLJ} a learning network can be `exported' to define a new
re-usable, stand-alone model type, whose fields are commonly the
component models (\texttt{tree} and \texttt{standardizer} in the
example above) and these can be mutated. For details, the reader is
referred to the `Composing Models' section of the \texttt{MLJ}
documentation \cite{MLJ}.

\subsection{Model stacking}

David Wolpert describes a rather general method for blending the
predictions of multiple models known as \textsl{model stacking}
\cite{Wolpert1992}. A basic two-layer stack consists of a number of
\textsl{base} learners and a single \textsl{adjudicating}
learner. When such a stack is called to make a prediction, the
individual predictions of the base learners are made the columns of a
feature table for the adjudicating learner, which then outputs the
final prediction. However, it is crucial to understand that the flow
of data \textit{during training} is not the same.

The base model predictions used to train the adjudicating model are
\textit{not} the predictions of the base learners fitted to all the
training data. Rather, to prevent the adjudicator giving too much
weight to the base learners with low training error, the input data is
first split into a number of folds (as in cross-validation). A base
learner is then trained on each fold complement individually, and
corresponding predictions on the folds are spliced together to form a
full-length prediction called an \textsl{out-of-sample prediction}. It
is these out-of-sample predictions that are used to train the
adjudicating model.

For readability, we limit our stacking illustration to two base
learners, with three folds for the out-of-sample predictions.  Each
base learner will get three separate machines, for training on each
fold complement, and a fourth machine, trained on all the supplied
data, for use in the prediction flow. Then there is one more machine
for training the adjudicator. Our code snippet makes use of the
following functions, assumed to have been overloaded to admit nodes as
arguments:
\begin{itemize}
\item \texttt{folds(X, n)}: Return an \texttt{n}-tuple of vectors of
  indices, as if for use in \texttt{n}-fold cross-validation. For
  example, if \texttt{X} is a table with ten rows, then
  \texttt{folds(X, n)} returns \texttt{([1, 2, 3], [4, 5, 6], [7, 8,
    9, 10])}.
\item \texttt{restrict(X, folds, i)}: Return the restriction of data
  object \texttt{X} to the \texttt{i}th fold of \texttt{folds}.
\item \texttt{corestrict(X, folds, i)}: Return the restriction of
  data object \texttt{X} to the \textit{complement} of the
  \texttt{i}th fold of \texttt{folds},
\item \texttt{vcat}, \texttt{hcat} --- vertical and horizontal
  concatenation
\item \texttt{MLJBase.table} --- operation converting a matrix to a table
\end{itemize}
For concreteness, we suppose the base models are a gradient tree
booster and a support vector machine; the adjudicator is a random
forest:
\begin{minted}[fontsize=\small,frame=lines]{julia}
@load EvoTreesRegressor
@load SVMRegressor;
@load RandomForestRegressor pkg=DecisionTree

model1 = EvoTreesRegressor(nrounds=100)
model2 = SVMRegressor()
judge = RandomForestRegressor(n_trees=500)

X = source()
y = source()

f = folds(X, 3) # a node!

# for training model1 on each fold complement:
m11 = machine(model1, corestrict(X, f, 1), corestrict(y, f, 1))
m12 = machine(model1, corestrict(X, f, 2), corestrict(y, f, 2))
m13 = machine(model1, corestrict(X, f, 3), corestrict(y, f, 3))

# predictions for model1 on each fold:
y11 = predict(m11, restrict(X, f, 1));
y12 = predict(m12, restrict(X, f, 2));
y13 = predict(m13, restrict(X, f, 3));

# for training model2 on each fold complement:
m21 = machine(model2, corestrict(X, f, 1), corestrict(y, f, 1))
m22 = machine(model2, corestrict(X, f, 2), corestrict(y, f, 2))
m23 = machine(model2, corestrict(X, f, 3), corestrict(y, f, 3))

# predictions for model2 on each fold:
y21 = predict(m21, restrict(X, f, 1));
y22 = predict(m22, restrict(X, f, 2));
y23 = predict(m23, restrict(X, f, 3));

# the out-of-sample predictions for base learners:
y1_oos = vcat(y11, y12, y13);
y2_oos = vcat(y21, y22, y23);

# make out-of-sample predictions columns of a table:
X_oos = MLJ.table(hcat(y1_oos, y2_oos))

# for training the adjudicator:
m_judge = machine(judge, X_oos, y)

# for training the base models on all available data:
m1 = machine(model1, X, y)
m2 = machine(model2, X, y)

# for processing *new* input features for feeding to the adjudicator:
y1 = predict(m1, X);
y2 = predict(m2, X)

# assembling the base-learner predictions on the new inputs:
X_judge = MLJ.table(hcat(y1, y2))

# for outputing the final prediction on new inputs:
yhat = predict(m_judge, X_judge)

\end{minted}
For complete MLJ code for this example, see the Stacking tutorial at
\cite{MLJTutorials}.

\end{document}